\pdfoutput=1

\documentclass[11pt]{article}

\usepackage[]{EMNLP2022}

\usepackage{times}
\usepackage{latexsym}

\usepackage[T1]{fontenc}

\usepackage[utf8]{inputenc}

\usepackage{microtype}

\usepackage{inconsolata}

\usepackage{hyperref}
\usepackage{url}
\usepackage{booktabs}
\usepackage{multirow}
\usepackage{amssymb}
\usepackage{pifont}
\usepackage{tabularx}
\newcolumntype{Y}{>{\centering\arraybackslash}X}
\usepackage{adjustbox}
\usepackage{amsmath,amsfonts,bm}
\usepackage{lipsum}
\usepackage{comment}
\usepackage{algorithm2e}
\usepackage{subcaption}
\usepackage{diagbox}
\usepackage{cleveref}
\usepackage{longtable}

\definecolor{cadmiumgreen}{rgb}{0.0, 0.42, 0.24}
\definecolor{burgundy}{rgb}{0.5, 0.0, 0.13}
\definecolor{darkviolet}{rgb}{0.58, 0.0, 0.83}
\definecolor{princetonorange}{rgb}{1.0, 0.56, 0.0}
\definecolor{darkmagenta}{rgb}{0.55, 0.0, 0.55}
\definecolor{cornellred}{rgb}{0.7, 0.11, 0.11}

\newcommand{\cmark}{\ding{51}}
\newcommand{\xmark}{\ding{55}}

\DeclareMathOperator*{\argmax}{arg\,max}

\renewcommand{\paragraph}[1]{\vspace{0.2cm}\noindent\textbf{#1}}

\newcommand{\ours}{GeNER}

\newcommand{\tallor}{\textsc{TaLLoR}}
\newcommand{\roberta}{RoBERTa}
\newcommand{\biobert}{BioBERT}
\newcommand{\bond}{\textsc{BOND}}
\newcommand{\quip}{\textsc{QuIP}}

\newcommand{\densephrases}{DensePhrases}
\newcommand{\typetoken}{[\texttt{TYPE}]}

\newcommand{\partiallyused}{{\footnotesize \boldsymbol{$\triangle$}}}
\newcommand{\xmarkspace}{\xmark$^{~\,}$}

\newcommand{\outperformbond}{$^\star$}
\newcommand{\smallspace}{$^{\;\:}$}

\newcommand{\precision}{$\text{P}$}
\newcommand{\recall}{$\text{R}$}
\newcommand{\fscore}{$\text{F}\text{1}$}

\DeclareSymbolFont{extraup}{U}{zavm}{m}{n}
\DeclareMathSymbol{\varheart}{\mathalpha}{extraup}{86}
\DeclareMathSymbol{\vardiamond}{\mathalpha}{extraup}{87}

\title{Simple Questions Generate Named Entity Recognition Datasets}

\author{Hyunjae Kim$^1$ \quad Jaehyo Yoo$^1$ \quad Seunghyun Yoon$^2$ \quad Jinhyuk Lee$^{1*}$ \quad Jaewoo Kang$^{1,3}$\\
$^1$Korea University \quad $^2$Adobe Research \quad $^3$AIGEN Sciences \\
\texttt{\{hyunjae-kim,jaehyoyoo,jinhyuk\_lee,kangj\}@korea.ac.kr} \\ \texttt{syoon@adobe.com}
}

\begin{document}
\maketitle

\begin{abstract}

Recent named entity recognition (NER) models often rely on human-annotated datasets, requiring the significant engagement of professional knowledge on the target domain and entities.
This research introduces an \textit{ask-to-generate} approach that automatically generates NER datasets by asking questions in simple natural language to an open-domain question answering system (e.g., ``\textit{Which disease?''}).
Despite using fewer in-domain resources, our models, solely trained on the generated datasets, largely outperform strong low-resource models by an average F1 score of 19.4 for six popular NER benchmarks.
Furthermore, our models provide competitive performance with rich-resource models that additionally leverage in-domain dictionaries provided by domain experts.
In few-shot NER, we outperform the previous best model by an F1 score of 5.2 on three benchmarks and achieve new state-of-the-art performance.
The code and datasets are available at \href{https://github.com/dmis-lab/GeNER}{https://github.com/dmis-lab/GeNER}.

\end{abstract}

\newcommand\blfootnote[1]{%
  \begingroup
  \renewcommand\thefootnote{}\footnote{#1}%
  \addtocounter{footnote}{-1}%
  \endgroup
}
\blfootnote{\textsuperscript{*}JL currently works at Google Research. The collaboration started before he joined Google.}

\vspace{-5mm}

\section{Introduction}

Named entity recognition (NER) is the task of extracting named entities of specific types from text.
An NER dataset essentially reflects the need to extract specific entity types.
For instance, NCBI-disease~\citep{dougan2014ncbi} was created for extracting \textit{disease} entities from text.
Recent NER models have provided robust performance when trained on carefully designed human-annotated datasets~\cite{lample-etal-2016-neural, li-etal-2020-unified, lee2020biobert}.
However, suppose that we want to build an NER model for extracting \textit{bacteria} or other specific types for which human-annotated datasets are insufficient.
Whenever we extract such entity types, should we rely on professional knowledge to create new datasets?

Previous weakly supervised NER models~\citep{shang-etal-2018-learning,liang2020bond} tackled this problem using rich in-domain dictionaries (e.g., The Comparative Toxicogenomics Database) and unlabeled training sentences (i.e., in-domain sentences), where entities in the dictionary are used to annotate the training sentences.
However, these approaches easily fail in practice because in-domain dictionaries and sentences are often unavailable or expensive to construct for many entity types.
It will be challenging to build NER models for \textit{enzyme} or \textit{astronomical object} entities without expert-level knowledge required for building dictionaries and searching for a large number of in-domain sentences for annotation.

In this study, we introduce \ours, an automated dataset \textbf{Ge}neration framework for \textbf{NER}, which automatically constructs high-quality NER datasets.
In particular, concrete needs for NER are described using simple natural language questions such as ``\textit{Which} \typetoken\textit{?}'', where \typetoken~is substituted by the required entity type (e.g., ``\textit{Which disease?}'').
Such questions do not require professional knowledge of the target domain and allow even non-experts to easily build domain-specific NER datasets.
Using a phrase retrieval model designed for open-domain question answering (QA)~\cite{lee-etal-2021-learning-dense}, \ours~first retrieves candidate entities (i.e., phrases) and evidence sentences from a large-scale open-domain corpus (e.g., Wikipedia).
The retrieved entities form a pseudo-dictionary, which is used to annotate the evidence sentences to create the dataset.
We then train standard NER models on our generated dataset using a recent self-training method~\citep{liang2020bond}.
As shown in \Cref{tab:categorization}, this type of \textit{ask-to-generate} approach significantly reduces dependency on the in-domain resources while outperforming the strong low-resource model, \tallor~\citep{li-etal-2021-weakly}, and being comparable with the rich-resource model, \bond~\citep{liang2020bond}.

\begin{table}[t!]
\centering
\begin{adjustbox}{max width = \columnwidth}
\begin{tabular}{lccc}
\toprule
 \multirow{1}{*}{\textbf{Model}} & \begin{tabular}[c]{@{}c@{}}\textbf{Resource}\\$\mathbf{X}_\text{train}$ / $\mathbf{Y}_\text{train}$ / $\mathcal{V}$\end{tabular} & \textbf{\begin{tabular}[c]{@{}c@{}}Wikigold\end{tabular}} & \textbf{\begin{tabular}[c]{@{}c@{}}NCBI-\\disease\end{tabular}} \\
\midrule
\multicolumn{4}{l}{\textit{Rich-resource models (w/ training label or in-domain dict.)}} \\
\midrule
 \begin{tabular}[c]{@{}l@{}} Fully supervised \end{tabular} & \cmark~/~\cmark~/~\xmark & 86.8 & 88.6 \\
 \begin{tabular}[c]{@{}l@{}}\bond  \end{tabular} & \cmark~/~\xmarkspace/~\cmark & 59.8 & 71.4 \\
\midrule
\multicolumn{4}{l}{\textit{Low-resource models}} \\
\midrule
 \tallor & \cmark~/~\xmarkspace/$^{\,}$\partiallyused & 30.3 & 44.3 \\
 \ours~(\textbf{ours}) & \xmarkspace/~\xmarkspace/~\xmark & 72.5 & 67.9 \\
\bottomrule
\end{tabular}
\end{adjustbox}
\caption{
Comparison of existing approaches in NER.
Each method is categorized based on how much it relies on in-domain resources during training.
$\mathbf{X}_\text{train}$: (unlabeled) training sentences. $\mathbf{Y}_\text{train}$: human-annotated training labels. $\mathcal{V}$: in-domain dictionaries by domain experts.
In-domain resources are either fully used (\cmark), partially used (\partiallyused), or not used (\xmark).
While using the fewest in-domain resources, \ours~shows strong performance on various domains and entity types.
}
\label{tab:categorization}
\vspace{-3mm}
\end{table}

We demonstrate the effectiveness of \ours~using six popular NER benchmarks across four domains: news \cite{tjong-kim-sang-de-meulder-2003-introduction}, Wikipedia \cite{balasuriya-etal-2009-named}, Twitter \cite{strauss-etal-2016-results}, and biomedicine \cite{dougan2014ncbi,li2016biocreative, krallinger2015chemdner}.
Models solely trained on our generated datasets from \ours~significantly outperformed \tallor~on all benchmarks by an average \fscore~score of 19.4.
Although our models did not use rich in-domain dictionaries, they sometimes outperformed the previous best weakly supervised model, \bond~\citep{liang2020bond}, on two benchmarks.
Moreover, GeNER achieved new state-of-the-art results on three few-shot NER benchmarks, outperforming the previous best model \quip~\cite{jia-etal-2022-question} by an F1 score of 5.2.
Finally, we conducted extensive ablation studies and analyses to highlight important factors for low-resource NER.

Our contributions are summarized as follows: 
\begin{itemize}
    \item{
    To the best of our knowledge, \ours~is the first attempt to automatically generate NER datasets for various low-resource domains using a general-purpose QA system.
    }
    \item{
    \ours~significantly reduces the dependency on in-domain training resources required by previous weakly supervised models, such as human-annotated labels, in-domain dictionaries, and in-domain sentences.
    }
    \item{
    \ours~outperformed the strong baseline model \tallor~by an F1 score of 19.4 on six benchmarks. 
    In few-shot NER, \ours~outperformed the previous best model \quip~by an F1 score of 5.2 on three benchmarks, thereby achieving new state-of-the-art results.
    }
\end{itemize}

\section{Background}

\subsection{Named Entity Recognition}

NER aims to identify named entities of pre-defined types in text.
Let $\mathcal{D} = \{\mathbf{X}, \mathbf{Y}\}$ be a dataset, where $\mathbf{X} = \{x_n\}_{n=1}^{N}$ is the set of unlabeled sentences, $\mathbf{Y} = \{y_n\}_{n=1}^{N}$ is the set of corresponding token-level labels\footnote{We follow the BIO tagging~\cite{ramshaw-marcus-1995-text}.} for each sentence, and $N$ is the size of the dataset.
In supervised learning, $\mathcal{D}$ is split into $\mathcal{D}_\text{train}= \{\mathbf{X}_\text{train}, \mathbf{Y}_\text{train}\}$, $\mathcal{D}_\text{valid}= \{\mathbf{X}_\text{valid}, \mathbf{Y}_\text{valid}\}$, and $\mathcal{D}_\text{test}= \{\mathbf{X}_\text{test}, \mathbf{Y}_\text{test}\}$, which are then used to train NER models, select hyperparameters, and evaluate the models, respectively.

\paragraph{Weakly supervised NER}
Instead of using human-annotated labels $\mathbf{Y}_\text{train}$, weakly supervised NER models rely on in-domain dictionaries $\mathcal{V}$ built by domain experts~\cite{yang-etal-2018-distantly,shang-etal-2018-learning,cao-etal-2019-low,yang-katiyar-2020-simple,liang2020bond}.
In-domain dictionaries are used to generate weak labels $\hat{\mathbf{Y}}_\text{train}$ for (unlabeled) training sentences $\mathbf{X}_\text{train}$ by annotating any occurrences of named entities from the dictionary.
Models are then trained on $\hat{\mathcal{D}}_\text{train}= \{\mathbf{X}_\text{train}, \hat{\mathbf{Y}}_\text{train}\}$ and evaluated on $\mathcal{D}_\text{test}$.
Instead of relying on training resources such as $\mathbf{X}_\text{train}, \mathbf{Y}_\text{train}$, and $\mathcal{V}$, we propose to automatically generate a new dataset $\tilde{\mathcal{D}}_\text{train}= \{\tilde{\mathbf{X}}_\text{train}, \tilde{\mathbf{Y}}_\text{train}\}$ with minimal human effort by asking simple questions to the QA model.

\begin{figure*}[t]
\centering
\includegraphics[width=\textwidth]{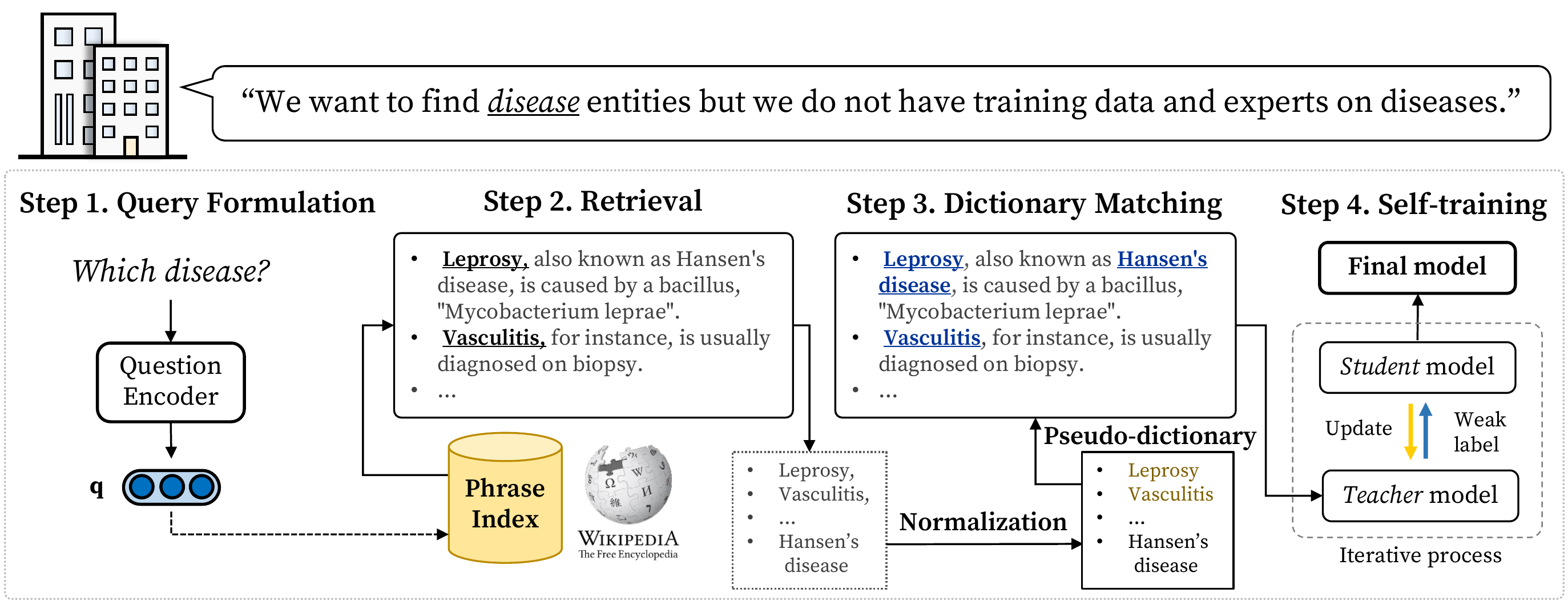}
\caption{
Overview of \ours.
Given the needs of extracting {disease} entities from text, \ours~automatically generates an NER dataset for the disease entities without resorting to human-annotated training data or domain experts.
(1)~Query formulation: the needs for disease NER are first formulated as simple natural language questions. (2)~Retrieval: we use an open-domain QA model to retrieve relevant phrases (i.e., entities) as well as sentences to annotate. (3)~Dictionary matching: retrieved sentences are annotated by the normalized phrases. (4)~Self-training: we train NER models purely on our generated dataset using self-training.
See~\Cref{sec:method} for more details.
}
\label{fig:intro}
\vspace{-3mm}
\end{figure*}

\subsection{Open-domain Question Answering}
Open-domain QA finds answers from a large-scale corpus that is not limited to specific domains~\citep{chen-etal-2017-reading}.
Among several other open-domain QA approaches, \citet{seo-etal-2019-real} proposed \textit{phrase retrieval}, which formulates question answering as retrieving pre-indexed phrases, significantly beneficial to scalability because the similarity search can be efficiently implemented.
Recent advancements in phrase retrieval~\citep{lee-etal-2021-learning-dense} have designed the retrieval purely with dense vectors as follows:
\begin{equation}
\label{equation:densephrases}
    \begin{aligned}
    &\mathbf{s} = E_{s}(s, \mathcal{W}),\quad \mathbf{q} = E_{q}(q), \\
    &s^* = \argmax_{s\in\mathcal{S}(\mathcal{W})}(\mathbf{s}^\top \mathbf{q}),
    \end{aligned}
\end{equation}
where $\mathcal{W}$ is a large-scale open-domain corpus; $\mathcal{S}(\mathcal{W})$ is the set of all phrases in $\mathcal{W}$; $q$ is the input question; and $E_{q}$ and $E_{s}$ are question and phrase encoders, respectively, that produce dense vectors $\mathbf{q}$ and $\mathbf{s}$, respectively.
The candidate answer $s^*$ can be retrieved along with an evidence document or sentence because of its extractive nature~\citep{lee-etal-2021-phrase}.
In this study, we used the phrase retrieval model DensePhrases~\citep{lee-etal-2021-learning-dense} as our open-domain QA model because of its strong open-domain QA accuracy and run-time efficiency while retrieving thousands of phrases and sentences to construct NER datasets.
We use evidence sentences that contain each $s^*$ as our training sentences $\tilde{\mathbf{X}}_\text{train}$ and leverage the retrieved phrases $s^*$ to generate the weak labels $\tilde{\mathbf{Y}}_\text{train}$.

\section{Method}
\label{sec:method}
This section describes \ours, which automatically generates NER datasets from scratch and provides NER models trained on these datasets.
\Cref{fig:intro} provides an overview of the \ours.

\subsection{Query Formulation}
\label{subsec:query_generation}

In \ours, we first formulate the need to recognize specific types of named entities as natural language questions. 
For instance, we ask, ``\textit{Which disease?}'' to extract {disease} entities.
Based on the ablation study in~\Cref{subsec:ablation_study}, we provide a template ``\textit{Which} \typetoken\textit{?},'' where \typetoken~is substituted by an entity type.
An input question $q$ is then encoded as question vector $\mathbf{q}$ as shown in Equation~\eqref{equation:densephrases}.
In real-world scenarios, practitioners have a small-sized entity-type ontology or list that defines entity types of interest; therefore, they can easily determine query terms (i.e., \typetoken) with minimal effort.

It should be noted that clear needs should be reflected in our questions.
For instance, rather than a single, broad, unspecified question, such as ``\textit{Which organization?}'', multiple distinct sub-questions such as ``\textit{Which sports team?}'' or ``\textit{Which company?}'', that represent the target entities specifically would be a better option (see \Cref{subsec:ablation_study}.).

\begin{table*}[t]
\centering
\footnotesize

\begin{adjustbox}{max width = 0.99\textwidth}

\begin{tabular}{llrrrrrr}
\toprule
\multirow{2}{*}{\textbf{Domain}} & \multirow{2}{*}{\textbf{Dataset (\# Types)}} & \multicolumn{2}{c}{\textbf{Training}} & \multicolumn{2}{c}{\textbf{Validation}} & \multicolumn{2}{c}{\textbf{Test}} \\
 &   & \textbf{\# Sents} & \textbf{\# Labels} & \textbf{\# Sents} & \textbf{\# Labels} & \textbf{\# Sents} & \textbf{\# Labels} \\
\midrule
News & CoNLL-2003 (3)  & 14,987 & 20,061 & 3,469 & 5,022  & 3,685 & 4,947  \\
\midrule
Wikipedia & Wikigold (3)  & 1,142 & 1,842 & 280 & 523 & 274 & 484  \\
\midrule
Twitter & WNUT-16 (9)  & 2,394 & 1,271 & 1,000 & 529 & 3,850 & 2,889  \\
\midrule
\multirow{3}{*}{Biomedicine} & NCBI-disease (1)  & 5,432 & 5,134 & 923 & 787 & 942 & 960 \\
& BC5CDR (2)  & 4,582 & 9,387 & 4,602 & 9,596 & 4,812 & 9,809  \\
& CHEMDNER (1) & 30,884 & 29,530 & 30,841 & 29,543 & 26,561 & 25,388 \\
\bottomrule
\end{tabular}

\end{adjustbox}

\caption{
Statistics of NER datasets.
\#~Types: number of entity types. 
\#~Sents: number of sentences.\
\#~Labels: number of entity-level annotations.
Note that we do not use any training sentences and labels from benchmarks but only the generated dataset from \ours~in the low-resource NER experiment (\Cref{subsec:low_resource}).
}
\label{tab:statistics}
\vspace{-3mm}
\end{table*}

\subsection{Retrieval}
Based on the question vector for each sub-question, we mine the unlabeled training sentences, $\tilde{\mathbf{X}}_\text{train}$, and generate the pseudo-dictionary, $\tilde{\mathcal{V}}$.
Given $L$ sub-questions for all entity types, we use \densephrases~to retrieve relevant phrases and evidence sentences.
While other domain-specific corpora, such as PubMed, can be used for domain-specific NER, we use Wikipedia\footnote{We use the 2018-12-20 Wikipedia snapshot.} as our corpus for retrieval because it covers many different domains, making \ours~generally applicable.

\paragraph{Normalization}
We retrieve the top phrases, $s^*$, for each sub-question and normalize them to refine their spans.
The set of normalized phrases comprises the pseudo-dictionary, $\tilde{\mathcal{V}}$.
Normalization rules are required because of different annotation guidelines of the datasets or the inherent characteristics of different entity types.
For instance, phrases containing the conjunction ``and'' should be split into two different phrases for CoNLL-2003, whereas this rule should not be applied to biomedical NER datasets because such phrases are considered a single \textit{composite mention}.
Furthermore, some entity types (e.g., song) can begin with the article ``the,'' whereas other entity types usually have no article in their names.
Therefore, we provide ten simple rules (\Cref{appendix:rule_description}) generally applicable to most entity types.
Practically, these rules can be determined by practitioners in a top-down manner or treated as hyperparameters.

\paragraph{Training sentences}
We obtain the top $k_l$ unique sentences for each sub-question that contains each phrase $s^*$.
We gather $k_1 + \dots + k_L$ sentences in total, which we consider as our unlabeled training sentences $\tilde{\mathbf{X}}_\text{train}$.
We do not have a hyperparameter for the size of the pseudo-dictionary, $\tilde{\mathcal{V}}$, and the size changes based on the number of unique sentences to be retrieved.\footnote{If we have more than two phrases contained in the same sentence, we retain all of them in $\tilde{\mathcal{V}}$.}

\subsection{Dictionary Matching}
This stage annotates the unlabeled sentences, $\tilde{\mathbf{X}}_\text{train}$, and generates $\tilde{\mathbf{Y}}_\text{train}$ using pseudo-dictionary $\tilde{\mathcal{V}}$ to prevent possible false negatives that might occur while annotating only initially retrieved phrases as entities.
For every phrase (or entity) in $\tilde{\mathcal{V}}$, every occurrence of the phrase in $\tilde{\mathbf{X}}_\text{train}$ is annotated to generate $\tilde{\mathbf{Y}}_\text{train}$.
If a phrase has more than two entity types, the resulting label ambiguity is dealt with using the probability of the phrase appearing with a particular entity type.
For instance, if ``Washington'' is retrieved three times for the location type and seven times for the person type, we annotate 30\% of all occurrences with location and the remaining 70\% with person.
Finally, we obtain the training data, $\tilde{\mathcal{D}}_\text{train} = \{\tilde{\mathbf{X}}_\text{train}, \tilde{\mathbf{Y}}_\text{train}\}$, which reflects our specific needs.

\begin{table*}[t]
\centering
\begin{subtable}{\textwidth}
\begin{adjustbox}{max width = 0.99\textwidth}
\begin{tabular}{lcrrrrrrrrr}
\toprule
\multirow{2}{*}{\textbf{Model}} & \textbf{Resource} & \multicolumn{3}{c}{\textbf{CoNLL-2003}} & \multicolumn{3}{c}{\textbf{Wikigold}}  & \multicolumn{3}{c}{\textbf{WNUT-16}}  \\
 & $\mathbf{X}_\text{train}$ / $\mathbf{Y}_\text{train}$ / $\mathcal{V}$ & \multicolumn{1}{c}{\textbf{\precision}} & \multicolumn{1}{c}{\textbf{\recall}} & \multicolumn{1}{c}{\textbf{\fscore}} & \multicolumn{1}{c}{\textbf{\precision}} & \multicolumn{1}{c}{\textbf{\recall}} & \multicolumn{1}{c}{\textbf{\fscore}} & \multicolumn{1}{c}{\textbf{\precision}} & \multicolumn{1}{c}{\textbf{\recall}} & \multicolumn{1}{c}{\textbf{\fscore}}  \\
 \midrule
\multicolumn{10}{l}{\textit{Rich-resource models (w/ training label or in-domain dict.)}} \\
\midrule
RoBERTa~\cite{liu2019roberta} & \cmark~/~\cmark~/~\xmark & {93.1}\smallspace & {93.9}\smallspace & {93.5}\smallspace & {86.5}\smallspace & {87.2}\smallspace & {86.8}\smallspace & {57.3}\smallspace & {60.8}\smallspace & {59.0}\smallspace  \\
\bond~\cite{liang2020bond} & \cmark~/~\xmarkspace/~\cmark & 80.6\smallspace & 82.4\smallspace & 81.5\smallspace & 58.2\smallspace & 61.5\smallspace & 59.8\smallspace & 47.0\smallspace & 48.4\smallspace & 47.7\smallspace  \\
\midrule
\multicolumn{10}{l}{\textit{Low-resource models}} \\
\midrule
Seed Entities & \xmarkspace/~\xmarkspace/$^{\,}$\partiallyused & \textbf{95.1}\outperformbond & 2.8\smallspace & 5.3\smallspace & \textbf{90.5}\outperformbond & 3.9\smallspace & 7.5\smallspace & \textbf{67.7}\outperformbond & 1.5\smallspace & 2.9\smallspace  \\
Neural Tagger & \cmark~/~\xmarkspace/$^{\,}$\partiallyused & 71.8\smallspace & 13.6\smallspace & 22.9\smallspace & 58.8\outperformbond & 4.1\smallspace & 7.7\smallspace & 0.5\smallspace & 7.4\smallspace & 1.0\smallspace  \\
Self-training & \cmark~/~\xmarkspace/$^{\,}$\partiallyused & 43.0\smallspace & 31.6\smallspace & 36.4\smallspace & 32.8\smallspace & 17.4\smallspace & 22.7\smallspace & 25.0\smallspace & 19.6\smallspace & 22.3\smallspace \\
\tallor$^\dagger$~\cite{li-etal-2021-weakly} & \cmark~/~\xmarkspace/$^{\,}$\partiallyused & 64.3\smallspace & 64.1\smallspace & 64.2\smallspace & -\smallspace & -\smallspace & -\smallspace & -\smallspace & -\smallspace & -\smallspace \\
\tallor~\cite{li-etal-2021-weakly} & \cmark~/~\xmarkspace/$^{\,}$\partiallyused & 59.3\smallspace & 58.4\smallspace & 60.2\smallspace & 35.0\smallspace & 26.8\smallspace & 30.3\smallspace & 32.0\smallspace & 23.7\smallspace & 27.2\smallspace \\
\ours~(\textbf{ours}) & \xmarkspace/~\xmarkspace/~\xmark & 73.1\smallspace & \textbf{69.0}\smallspace & \textbf{71.0}\smallspace & 65.8\outperformbond & \textbf{79.9}\outperformbond & \textbf{72.5}\outperformbond & 44.8\smallspace & \textbf{54.0}\outperformbond & \textbf{48.5}\outperformbond \\
\bottomrule
\end{tabular}
\end{adjustbox}
\label{tab:main_general}
\end{subtable}

\vspace*{0.4 cm}

\begin{subtable}{\textwidth}
\begin{adjustbox}{max width = 0.99\textwidth}

\begin{tabular}{lcrrrrrrrrr}
\toprule
\multirow{2}{*}{\textbf{Model}} & \textbf{Resource} & \multicolumn{3}{c}{\textbf{NCBI-disease}} & \multicolumn{3}{c}{\textbf{BC5CDR}} & \multicolumn{3}{c}{\textbf{CHEMDNER}} \\
 & $\mathbf{X}_\text{train}$ / $\mathbf{Y}_\text{train}$ / $\mathcal{V}$ &  \multicolumn{1}{c}{\textbf{\precision}} & \multicolumn{1}{c}{\textbf{\recall}} & \multicolumn{1}{c}{\textbf{\fscore}} & \multicolumn{1}{c}{\textbf{\precision}} & \multicolumn{1}{c}{\textbf{\recall}} & \multicolumn{1}{c}{\textbf{\fscore}} & \multicolumn{1}{c}{\textbf{\precision}} & \multicolumn{1}{c}{\textbf{\recall}} & \multicolumn{1}{c}{\textbf{\fscore}}  \\
\midrule
\multicolumn{10}{l}{\textit{Rich-resource models (w/ training label or in-domain dict.)}} \\
\midrule
\biobert~\cite{lee2020biobert} & \cmark~/~\cmark~/~\xmarkspace & 86.6\smallspace & 90.5\smallspace & 88.5\smallspace & 86.7\smallspace & 90.5\smallspace & 88.6\smallspace & 91.4\smallspace & 91.1\smallspace & 91.2\smallspace \\
\bond~\cite{liang2020bond} & \cmark~/~\xmarkspace/~\cmark & 87.5\smallspace & 60.3\smallspace & 71.4\smallspace & 81.0\smallspace & 80.3\smallspace & 80.6\smallspace & -\smallspace & -\smallspace & -\smallspace \\
\midrule
\multicolumn{10}{l}{\textit{Low-resource models}} \\
\midrule
Seed Entities & \xmarkspace/~\xmarkspace/$^{\,}$\partiallyused & \textbf{88.8}\outperformbond & 10.7\smallspace & 19.1\smallspace & {\textbf{95.7}}\outperformbond & 3.6\smallspace & 6.9\smallspace & \textbf{93.7}\smallspace & 12.2\smallspace & 21.5\smallspace \\
Neural Tagger & \cmark~/~\xmarkspace/$^{\,}$\partiallyused & 75.2\smallspace & 24.9\smallspace & 37.4\smallspace & {93.1}\outperformbond & 9.7\smallspace & 17.6\smallspace & 74.8\smallspace & 21.6\smallspace & 33.5\smallspace  \\
Self-training & \cmark~/~\xmarkspace/$^{\,}$\partiallyused & 67.5\smallspace & 35.1\smallspace & 46.2\smallspace & 73.3\smallspace & 12.7\smallspace & 21.6\smallspace & 41.2\smallspace & 44.7\smallspace & 42.9\smallspace \\
\tallor$^\dagger$~\cite{li-etal-2021-weakly} & \cmark~/~\xmarkspace/$^{\,}$\partiallyused & -\smallspace & -\smallspace & -\smallspace & 66.5\smallspace & 66.9\smallspace & 66.7\smallspace & 63.0\smallspace & 60.2\smallspace & 61.6\smallspace \\
\tallor~\cite{li-etal-2021-weakly} & \cmark~/~\xmarkspace/$^{\,}$\partiallyused & 61.5\smallspace & 34.7\smallspace & 44.3\smallspace & 65.6\smallspace & 56.8\smallspace & 61.9\smallspace & 61.6\smallspace & 51.5\smallspace & 56.1\smallspace \\
\ours~(\textbf{ours}) & \xmarkspace/~\xmarkspace/~\xmark & {75.0}\smallspace & {\textbf{62.1}}\outperformbond & \textbf{67.9}\smallspace & 71.9\smallspace & \textbf{76.8}\smallspace & \textbf{74.3}\smallspace & {60.3}\smallspace & \textbf{64.4}\smallspace & \textbf{62.3}\smallspace \\
\bottomrule
\end{tabular}
\end{adjustbox}
\label{tab:main_biomedical}
\end{subtable}

\caption{
Performance of NER models on six datasets.
$\mathbf{X}_\text{train}$: (unlabeled) training sentences. $\mathbf{Y}_\text{train}$: human-annotated training labels. $\mathcal{V}$: in-domain dictionaries by domain experts.
In-domain resources are either fully used (\cmark), partially used (\partiallyused), or not used (\xmark).
$^\dagger$: utilizes n-gram statistics from the test set ($\mathbf{Y}_\text{test}$).
Among low-resource models, best scores are marked in boldface and scores higher than that of \bond~are denoted as \outperformbond.
}
\label{tab:main_table}
\vspace{-3mm}
\end{table*}

\subsection{Self-training}
As the weak labels generated by dictionary matching are often noisy and incomplete, directly training NER models using $\tilde{\mathcal{D}}_\text{train}$ is not optimal.
Therefore, we train our models using a current self-training method~\cite{liang2020bond} to mitigate noise.
First, we initialize \textit{teacher} and \textit{student} models using the same pre-trained weights of RoBERTa or BioBERT.
We train the \textit{teacher} model with the generated $\tilde{\mathcal{D}}_\text{train}$ for $T_\text{begin}$ steps.
The teacher model then re-annotates $\tilde{\mathbf{X}}_\text{train}$ and the \textit{student} model is trained on the re-annotated corpus.
The teacher model is replaced with the student model for every $T_\text{update}$ step, where $T_\text{update}$ denotes the update period.
This process is iterated until the maximum epoch is reached.
We use the student model with the best validation F1 score during the process as the final NER model.
It should be noted that GeNER is a model-agnostic framework; therefore, other recent techniques~\cite{liu-etal-2021-noisy-labeled,meng-etal-2021-distantly} can be adopted to correct mislabeling.

\section{Experiments}
We evaluated GeNER in two scenarios wherein training resources are scarce: low-resource NER (\Cref{subsec:low_resource}) and few-shot NER (\Cref{subsec:few_shot}).
We used entity-level precision (\precision), recall (\recall), and \fscore~score (\fscore) as the evaluation metrics.

\paragraph{Query term selection}
Evaluating our models requires investigating the needs inherent in the benchmark datasets.
In most NER benchmarks, entity types are coarsely defined and can be classified into many subtypes.
These subtypes are often defined differently based on the datasets.
For instance, the organization type of CoNLL-2003 mostly includes sports teams and companies, but that of Wikigold additionally covers bands.
Therefore, to understand the needs of each NER dataset, we sampled 100 examples from each validation set and analyzed them to formulate adequate sub-questions.
For instance, we used nine sub-questions for CoNLL-2003. 
All sub-questions for each dataset are presented in \Cref{tab:discrete_config} (Appendix) owing to space limitations.

\paragraph{NER models}
For the teacher and student models, we used \roberta~\cite{liu2019roberta} with a simple linear classifier for the token-level prediction in most experiments.
For biomedical-domain datasets, \biobert~\cite{lee2020biobert} was used as the backbone language model.

\subsection{Low-resource NER}
\label{subsec:low_resource}
This experiment assumed that human-annotated training labels and dictionaries are not available.
Following \citet{li-etal-2021-weakly}, we used the validation sets to search for the best hyperparameters and model checkpoints.

\paragraph{Datasets}
We used six popular NER benchmarks across four domains: \footnote{Following \citet{li-etal-2021-weakly}, we exclude the \textit{miscellaneous} and \textit{others} types because the needs for entities are not clarified.}
(1) \textbf{CoNLL-2003} \citep{tjong-kim-sang-de-meulder-2003-introduction} comprises Reuters news articles with three entity types: person, location, and organization.
(2) \textbf{Wikigold} \citep{balasuriya-etal-2009-named} has the same entity types as CoNLL-2003, but their subcategories are drastically different because of domain differences.
In addition, its size is relatively small compared to that of others.
(3) \textbf{WNUT-16} \cite{strauss-etal-2016-results} comprises nine entity types annotated on tweets, such as TV show, movie, and musician.
(4) \textbf{NCBI-disease} \citep{dougan2014ncbi} is a corpus of 793 PubMed abstracts with manually annotated disease entities.
(5) \textbf{BC5CDR} \citep{li2016biocreative} comprises 1,500 manually annotated PubMed abstracts with disease and chemical entities.
(6) \textbf{CHEMDNER} \citep{krallinger2015chemdner} is a corpus of 10,000 PubMed abstracts with manually annotated chemical entities; it is the largest corpus in our experiments.
\Cref{tab:statistics} lists the benchmark statistics.

\paragraph{Baselines}
\label{subsec:baseline_models}
We compared \ours~with other \textit{low-resource} models that do not use a full-size in-domain dictionary.
Among the previous low-resource models, \tallor~\cite{li-etal-2021-weakly} uses the least amount of in-domain resources: unlabeled training sentences $\mathbf{X}_\text{train}$ and the set of seed entities, which is a small dictionary $\mathcal{V}$ that contains 20-60 manually selected (i.e., $\mathcal{V}=$ \partiallyused).
In addition to \tallor, we provide baselines that use similar in-domain resources as \tallor, i.e., Seed Entities, Neural Tagger, and Self-training.
More details of each baseline are presented in Appendix~\ref{appendix:low-baseline}.
Additionally, we also report the performance of \textit{rich-resource} models that have access to either human-annotated training labels $\mathbf{Y}_\text{train}$ or rich in-domain dictionaries $\mathcal{V}$ constructed by domain experts.
This type of model includes fully supervised \roberta~\cite{liu2019roberta}, \biobert~\cite{lee2020biobert}, and the previous best weakly supervised model \bond~\cite{liang2020bond}.

\paragraph{Results}
As shown in Table~\ref{tab:main_table}, despite using fewer in-domain resources, \ours~outperformed all low-resource models in terms of \fscore~score.
\ours~significantly outperformed the strongest low-resource model, \tallor, by an average \fscore~score of 19.4 (macro averaged over six datasets).
Although \ours~used unlabeled sentences retrieved from Wikipedia, it delivered excellent performance on noisy user-generated text (i.e., WNUT-16) and scientific literature (i.e., biomedical-domain datasets), thereby indicating that our model is applicable to various text genres and domains.

Interestingly, \ours~even outperformed \bond~on Wikigold and WNUT-16 by F1 scores of 12.3 and 0.8, respectively.\footnote{It is difficult to train \bond~on CHEMDNER owing to the lack of rich dictionaries, which indicates the limitation of previous weakly supervised methods.}
These results indicate that our approach, which automatically generates a pseudo-dictionary, is promising and can be comparable to methods that use an expert-provided dictionary.

\begin{table}[t!]
\centering
\footnotesize
\begin{adjustbox}{max width =\columnwidth}
\begin{tabular}{lccc}
\toprule
\textbf{Model} & \textbf{\begin{tabular}[c]{@{}c@{}}CoNLL\\-2003\end{tabular}} & \textbf{Wikigold} & \textbf{BC5CDR}  \\
\midrule
Supervised & 53.5$^\ddagger$ & 47.0$^\ddagger$ & 55.0\\
\quad+ NSP  & 61.4$^\ddagger$ & 64.0$^\ddagger$ & - \\
\qquad+ Self-training & 65.4$^\ddagger$ & 68.4$^\ddagger$ & -   \\
\quip~(standard) & 70.0$^\ddagger$  & 67.6\smallspace & 61.8   \\
\quip~(prompt) & 74.0$^\ddagger$  & 70.6\smallspace & 65.7  \\
\midrule
\ours & 71.0\smallspace & 72.5\smallspace & 74.3 \\
\quad+ Fine-tuning   &  \textbf{75.0}\smallspace & \textbf{73.3}\smallspace & \textbf{77.7}  \\
\bottomrule
\end{tabular}
\end{adjustbox}
\caption{
Performance of few-shot NER models on three NER datasets.
\fscore~score is reported.
$^\ddagger$ indicates that scores are from~\citet{huang-etal-2021-shot} and \citet{jia-etal-2022-question}.
}
\label{tab:few_shot_ner}
\vspace{-3mm}
\end{table}

\subsection{Few-shot NER}
\label{subsec:few_shot}

Few-shot NER is another approach for addressing low-resource problems.
Unlike the use of the entire training dataset, few-shot NER models use a smaller number of training sentences and their labels (that is, $\mathbf{X}_\text{train}=$ \partiallyused and $\mathbf{Y}_\text{train}=$ \partiallyused).
We evaluated (1) \textbf{\ours}, which does not use even a small number of human-annotated examples, and (2) \textbf{\ours~+ Fine-tuning}, which is initialized by the best checkpoint of \ours~and then fine-tuned with the token-level prediction objective using the few-shot training examples until it converges. 

\paragraph{Settings}
We compared \ours~with the methods of \citet{huang-etal-2021-shot}~and \citet{jia-etal-2022-question} using three NER datasets used in the prior works: CoNLL-2003, Wikigold, and BC5CDR.
Details of the baseline models are presented in Appendix~\ref{appendix:few_shot}.
In total, 20 training examples were provided for CoNLL-2003 and Wikigold, whereas only 10 were provided for BC5CDR.
All the results were averaged over five different sampled datasets with the same number of examples.
Unlike the low-resource NER experiment, the \textit{miscellaneous} type was included in the experiment for a fair comparison with the baselines.

\paragraph{Results}
Table~\ref{tab:few_shot_ner} shows the performance of the few-shot NER models and \ours.
GeNER outperformed the previous best model, \quip~\citep{jia-etal-2022-question}, on two datasets, even before its fine-tuning.
When fine-tuned on the same set of few-shot examples, \ours~achieved a new state-of-the-art performance on all datasets.

\section{Analysis}
\label{sec:analysis}

\subsection{Ablation Study}
\label{subsec:ablation_study}

\paragraph{Question templates}
\label{subsec:query_template_selection}
We tested five different question templates in \ours~and compared them in terms of their phrase retrieval quality and final NER performance.
To measure the retrieval quality, we manually checked how many of the top 100 phrases for each sub-question were entities of correct types and computed the precision (P@100).
Furthermore, we measured the number of unique phrases in the top-100 retrievals (i.e., Diversity).
Table \ref{tab:query_template_search} shows that ``\textit{Which} \typetoken\textit{?}'' has the highest P@100 and ``\typetoken'' has the best diversity.
Although the diversity measure correlates well with the performance while retrieving $k_l=100$ sentences for each sub-question, retrieving a larger number of sentences ($k_l=5,000$) mitigates the low-diversity problem and provides the best overall performance.
\begin{table}[t]
\centering
\footnotesize
\begin{adjustbox}{max width = \columnwidth}
\begin{tabular}{lcccc}
\toprule
\multirow{2}{*}{\textbf{Template}} & \multirow{2}{*}{\textbf{P@100}} & \multirow{2}{*}{\textbf{Diversity}} & \multicolumn{2}{c}{\textbf{\fscore~Score}} \\
& & & $k_l=100$ & $5,000$ \\
\midrule
\multirow{1}{*}{Which \typetoken?} & \textbf{97.4} & 44.6 & 52.3 & \textbf{72.7}  \\
\midrule
\multirow{1}{*}{list of \typetoken} & 79.4 & 56.3 & 53.6 & 72.1  \\
\multirow{1}{*}{example of \typetoken} & 66.4 & 50.9 & 49.7 & 57.9  \\
\multirow{1}{*}{What \typetoken?} & 90.7 & 48.7 & 53.3 & 61.0 \\
\multirow{1}{*}{\typetoken} & 69.6 & \textbf{58.9} & \textbf{56.1} & 67.4 \\
\bottomrule
\end{tabular}
\end{adjustbox}
\caption{
Retrieval quality and NER performance of different question templates.
P@100 and diversity are macro-averaged over different types, and \fscore~score on the CoNLL-2003 validation set is reported.
}
\label{tab:query_template_search}
\end{table}

\begin{table}[t!]
\centering
\footnotesize
\begin{adjustbox}{max width = \columnwidth}
\begin{tabular}{lcc}
\toprule \relax
\typetoken & \textbf{CoNLL-2003} & \textbf{Wikigold}  \\
\midrule
organization & 27.3 & 35.8 \\
\midrule
sports team & 49.9 & 46.8 \\
\quad+ company & 53.3 & 57.2 \\
\qquad+ band & \textbf{55.3} & \textbf{60.7} \\
\bottomrule
\end{tabular}
\end{adjustbox}
\caption{
Performance of \ours~with different sets of sub-questions for the organization type on the CoNLL-2003 and Wikigold validation sets.
F1 score on the organization type is reported.
Each \typetoken~is used with the question template ``\textit{Which} \typetoken\textit{?}''.
}
\label{tab:query_ablation}
\vspace{-3mm}
\end{table}

\paragraph{Effect of sub-questions}
\ours~uses sub-questions to better reflect the needs inherent in each NER dataset.
In \Cref{tab:query_ablation}, we report the performance of \ours~on the CoNLL-2003 and Wikigold validation sets with different sets of sub-questions.
Using multiple sub-questions provides better performance while being more explicit about the needs than the performance while using only the ``\textit{Which organization?}'' question. 
Interestingly, although CoNLL-2003 does not contain many band names unlike Wikigold, both datasets benefit from using ``band'' as an additional sub-question, implying that their context may help generalize to other organizational entities.

\begin{figure}[t]
\centering
\includegraphics[width=0.95\columnwidth]{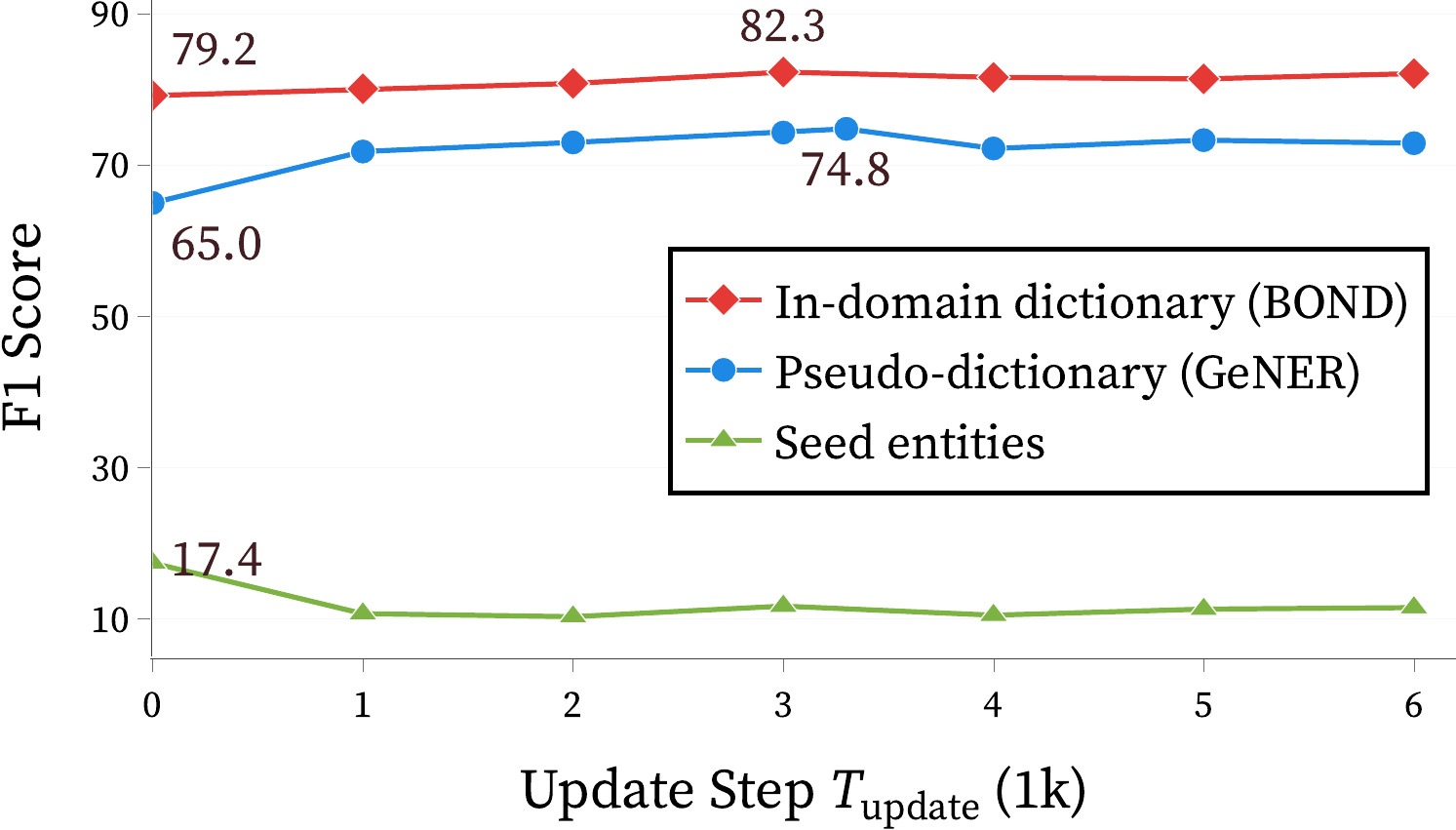}
\caption{
Effect of self-training with three different dictionaries from \tallor, \bond, and \ours. F1 score on the BC5CDR validation set is reported.
Best scores for each dictionary are labeled.
}
\label{fig:self_training}
\vspace{-3mm}
\end{figure}

\paragraph{Effect of self-training}
\Cref{fig:self_training} illustrates the effect of self-training with three different dictionaries: seed entities from \tallor, an in-domain dictionary from \bond, and a pseudo-dictionary from \ours.\footnote{The size of the dictionary from \bond~is more than 300k, and ours is 15k. The set of seed entities comprises 20 entities.}
The performance at step=$0$ represents the performance immediately after the teacher model is first initialized.
Although our pseudo-dictionary is initially incomplete compared with the in-domain dictionary, self-training largely closes the gap, which does not occur for the seed entities.

\begin{figure*}[t]
\centering

\begin{subfigure}{.3\textwidth}
  \centering
  \includegraphics[width=.9\linewidth]{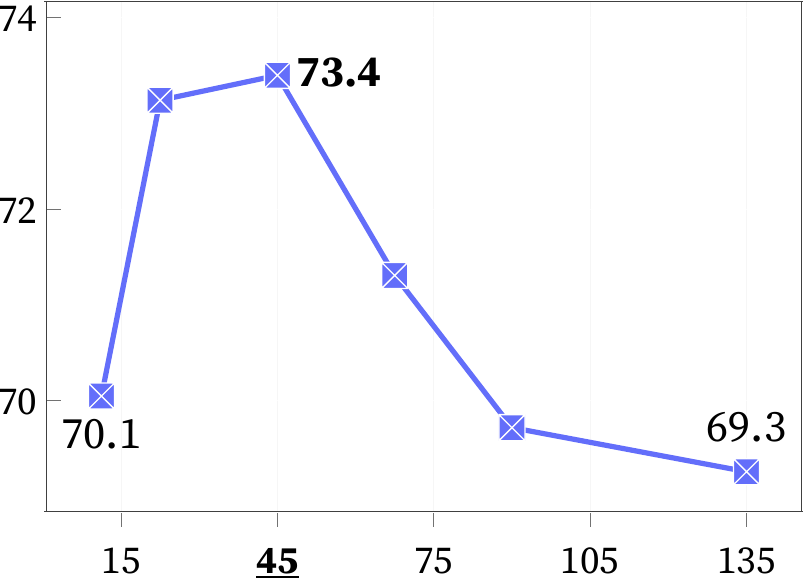} 
  \caption{CoNLL-2003}
  \label{fig:sub-first}
\end{subfigure}
\begin{subfigure}{.3\textwidth}
  \centering
  \includegraphics[width=.9\linewidth]{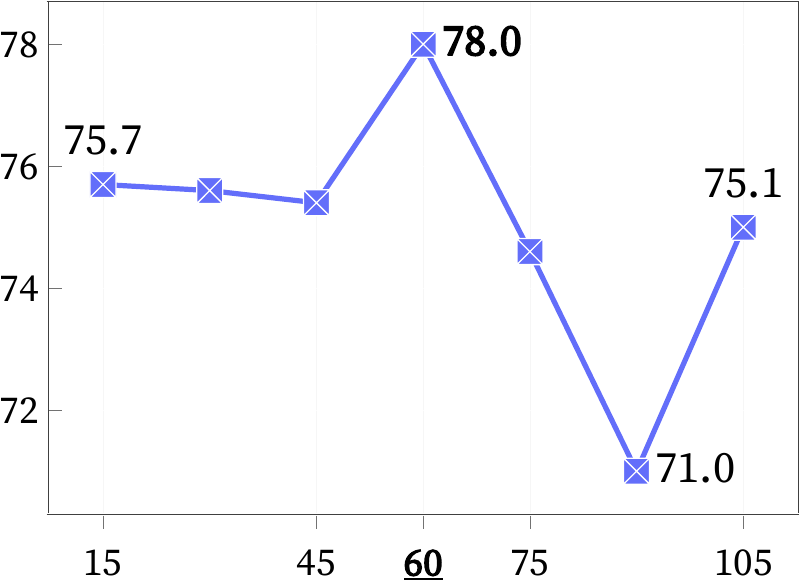}  
  \caption{Wikigold}
  \label{fig:sub-second}
\end{subfigure}
\begin{subfigure}{.3\textwidth}
  \centering
  \includegraphics[width=.9\linewidth]{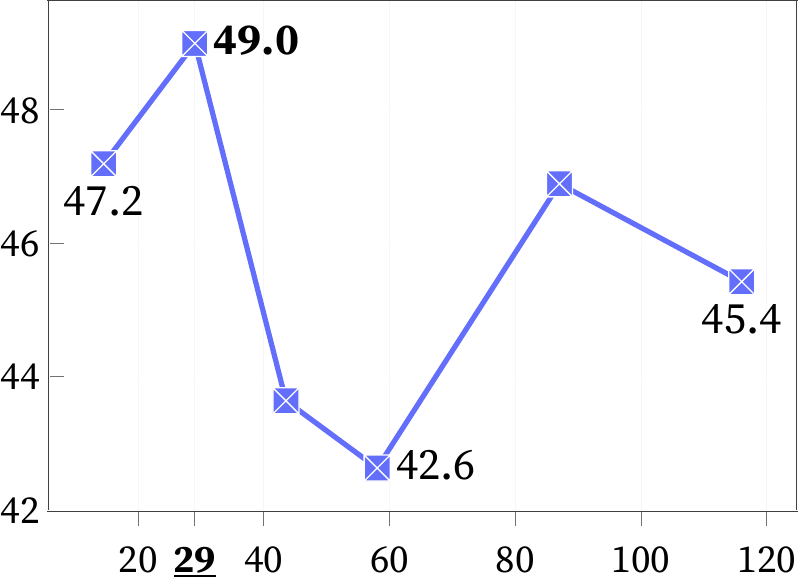}
  \caption{WNUT-16}
  \label{fig:sub-third}
\end{subfigure}

\vspace{+3mm}

\begin{subfigure}{.3\textwidth}
  \centering
  \includegraphics[width=.9\linewidth]{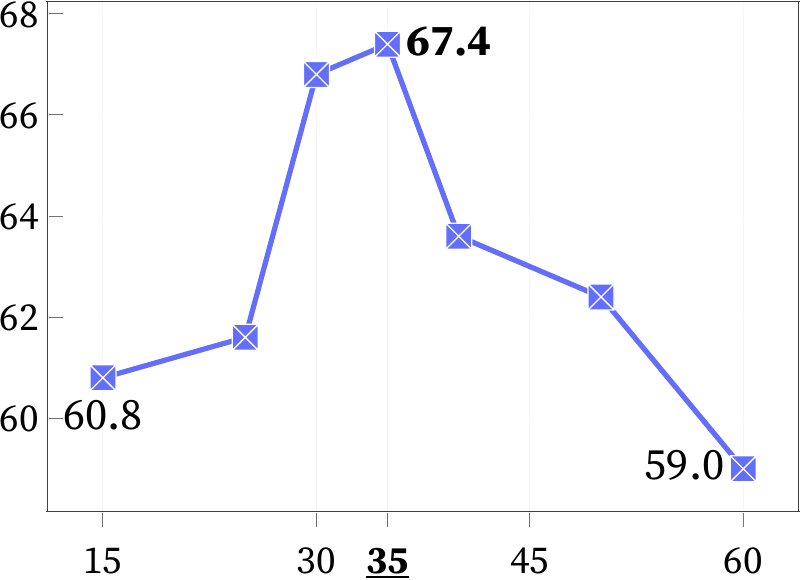}
  \caption{NCBI-disease}
  \label{fig:sub-third}
\end{subfigure}
\begin{subfigure}{.3\textwidth}
  \centering
  \includegraphics[width=.9\linewidth]{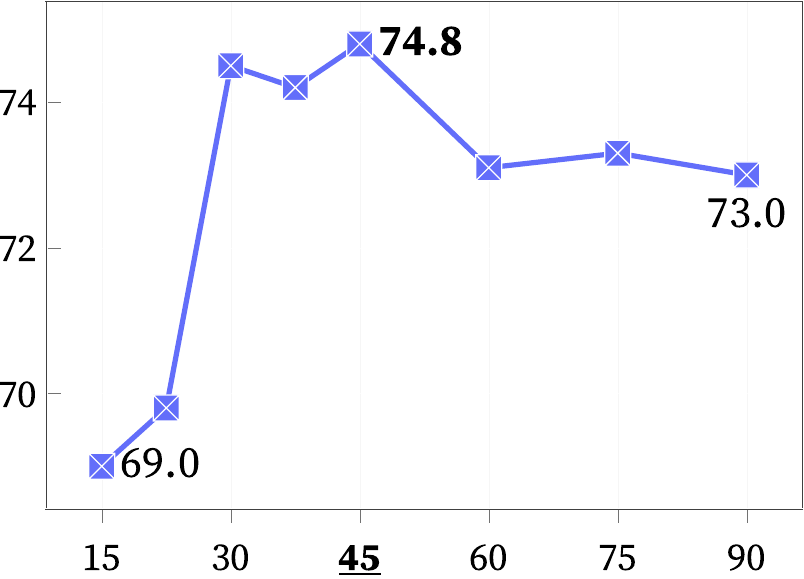}  
  \caption{BC5CDR}
  \label{fig:sub-fifth}
\end{subfigure}
\begin{subfigure}{.3\textwidth}
  \centering
  \includegraphics[width=.9\linewidth]{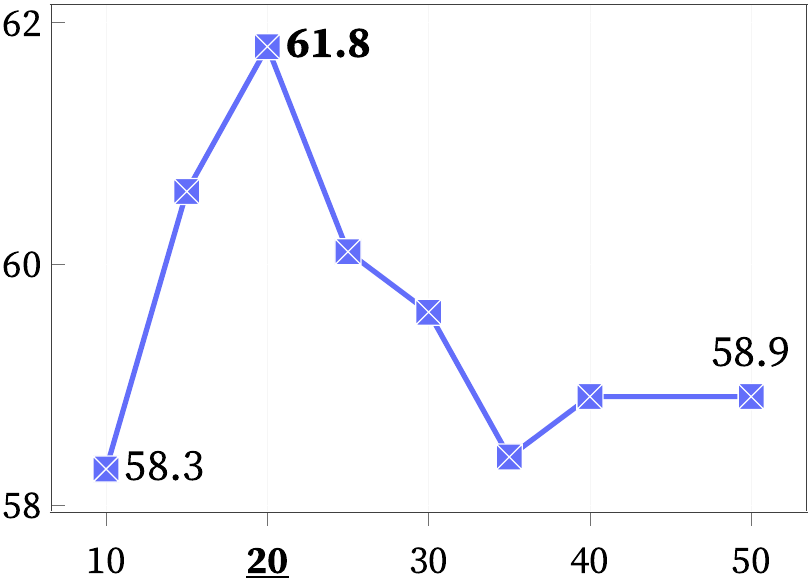}  
  \caption{CHEMDNER}
  \label{fig:sub-sixth}
\end{subfigure}
\caption{Performance of \ours~with different number of retrieved sentences on six datasets. The X and Y axes of the graphs indicate the total number of sentences (1k) and the F1 scores on the validation sets, respectively.}
\label{fig:top_k_all}
\end{figure*}
\paragraph{Effect of number of sentences}
\Cref{fig:top_k_all} shows how the performance changes when the total number of retrieved sentences for each dataset is increased.
The performance tends to improve at first but degrades after the highest is reached, indicating that simply increasing the number of entities does not help.
We suspect that the number of phrases with incorrect types may be increasing, causing performance degradation.

\subsection{Qualitative Analysis}
\label{subsec:visualization}

\begin{figure}[t]
\centering
\includegraphics[width=\columnwidth]{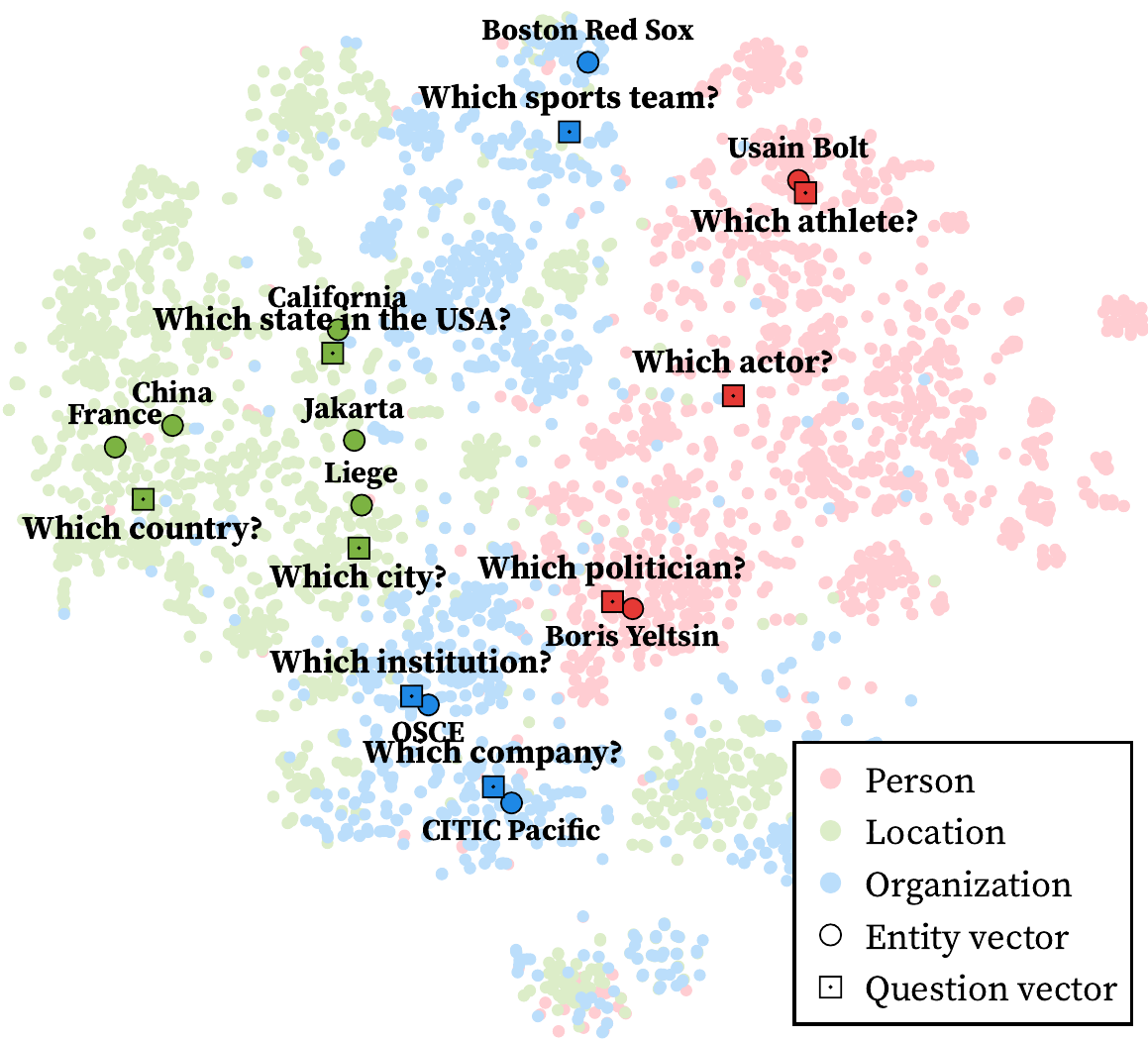}
\caption{
Visualization of \ours~question vectors and entities in the CoNLL-2003 validation set.
Vectors are visualized with t-SNE.
}
\label{fig:visualization}
\end{figure}

\begin{table*}[t!]
\centering
\footnotesize
\begin{adjustbox}{max width = 0.99\textwidth}

\begin{tabular}{l}
\toprule

\tallor~(\textbf{entities} and their context for each rule) \\
\midrule
Rule: \texttt{POStag}=``\texttt{NOUN}'' $\wedge$ \texttt{PostNgram}=``\text{attack}'' \\ 
\relax[1] Acute \textbf{hepatitis} attack after exposure to telithromycin. \\
\relax[2] This is not consistent with a CNS origin of \textbf{migraine} attack. \\ 
\midrule
\begin{tabular}[c]{@{}l@{}} Rule: \texttt{PreNgram}=``\text{in patients with}'' $\wedge$ \texttt{PostNgram}=``\text{'s disease}'' \\
 \relax[1] $\dots$ an increased mortality in patients with \textbf{Parkinson's disease} (PD) $\dots$ \end{tabular} \\
 \relax[2] $\dots$ in the treatment of psychosis and disruptive behaviors in patients with \textbf{Alzheimer's disease}. \\

\bottomrule
\\
\toprule
\ours~(\textbf{entities} and their context for ``\textit{Which disease?}'')\\
\midrule
\relax[1] \textbf{Leprosy} has affected humanity for thousands of years. \\
\relax[2] \textbf{Heart disease} is one of the leading causes of death in the world. \\
\relax[3] During this war an outbreak of \textbf{syphilis} occurred among the French troops. \\
\relax[4] $\dots$, \textbf{typhus} being at once the most contagious and the most preventable of diseases, $\dots$ \\
\relax[5] When \textbf{syphilis} was first definitely recorded in Europe in 1495, its pustules often $\dots$ \\
\bottomrule
\end{tabular}

\end{adjustbox}
\caption{Comparison of extracted entities and their context from \tallor~\citep{li-etal-2021-weakly} and \ours~(ours) for disease NER. While \tallor~relies on explicit rules based on POS tags or n-grams, \ours~discovers named entities more implicitly, which appear in more diverse context. Note that the context of \tallor~is from the BC5CDR training set (sentences from PubMed) while that of \ours~is mined from Wikipedia.}
\label{tab:compare_with_tallor}
\vspace{-3mm}
\end{table*}

\paragraph{Visualization}
The top phrases of \ours~are retrieved based on the similarity score between the phrase (i.e., entity) vectors and our question vector.
To understand how \ours~works, we visualized question vectors $\mathbf{q}$ used by \ours~and the entity vectors computed from the CoNLL-2003 validation set.
The question vectors are encoded from the question encoder of \densephrases, whereas its phrase encoder is used to compute the entity vectors of the annotated entities in CoNLL-2003.
From Figure \ref{fig:visualization}, it can be observed that the entities in the validation set are well separated based on their entity types, indicating that the phrase encoder of \densephrases~provides high-quality entity representations for NER.
In addition, we observe that our question vectors cover different groups of entity vectors, which eventually retrieve entities of correct types.

\paragraph{Context diversity}
Dense representations of text can capture subtle semantic relationships between the context and question~\citep{karpukhin-etal-2020-dense,lee-etal-2021-learning-dense}; therefore, our simple questions often retrieve sentences with diverse contexts.
We found that almost half of the retrieved sentences for ``\textit{Which disease?}'',  do not contain ``disease'' in their context.
As shown in \Cref{tab:compare_with_tallor}, our retrieved sentences have a considerably diverse context than sentences from the rule-based model~\citep{li-etal-2021-weakly}.

\begin{table}[t]
\centering
\footnotesize
\begin{adjustbox}{max width =\columnwidth}

\begin{tabular}{lr}
\toprule
\textbf{Step} & \textbf{Time}  \\
\midrule
1. Query Formulation & \hfill 0.7s \\
2. Retrieval & \hfill 1m 22.5s  \\
3. Dictionary Matching & \hfill 9m 16.3s  \\
4. Self-training & \hfill 29m$^{\,\,\,}$ 4.2s \\
\midrule
Total Time  & \hfill 39m 43.7s  \\ 
\bottomrule
\end{tabular}
\end{adjustbox}
\caption{
Complexity analysis of \ours~on CoNLL-2003. For the self-training step, time taken for finding the best model on each validation set is reported.
Time per each step is measured on a server with Intel Xeon(R) Silver 4210 CPU @ 2.20GHz and a single 24GB GPU (RTX 3090).
}
\label{tab:complexity}
\vspace{-3mm}
\end{table}

\subsection{Complexity Analysis}
From automatic dataset generation to model training, \ours~is highly efficient. 
The dataset generation steps (Steps 1, 2, and 3) mostly required approximately 10 min in total, while the self-training step (Step~4) required approximately 30 min.

\section{Discussion and Conclusion}
\ours~is the first attempt to automatically generate NER datasets using a general-purpose QA system.
GeNER, while using fewer in-domain resources, largely outperformed existing low-resource models on all six benchmarks, sometimes even outperforming the rich-resource model BOND. 
GeNER achieved a new state-of-the-art performance on three benchmarks upon evaluating the few-shot setting.
Our code and datasets have been made publicly available to facilitate further research.
We discuss some of the important aspects of \ours~that have not been explored in depth and provide possible future directions.

\paragraph{Better QA models}
\ours~is a model-agnostic framework; therefore, we can employ stronger open-domain QA models that often rely on the retriever-reader approach~\citep{fajcik-etal-2021-r2-d2}.
However, because of the large number of phrases that are required to be retrieved (e.g., 5,000), it is highly convenient to use phrase retrieval models with better run-time efficiency and strong accuracy.\footnote{Our preliminary experiments of using DPR~\citep{karpukhin-etal-2020-dense} showed much lower retrieval performance (P@100) and slower inference speed. It is also difficult to scale DPR to extract more than a hundred phrases.}
Whether the advancement of open-domain QA models can translate to the improvement of \ours~is an interesting research direction.

\paragraph{Other applications}
Other applications of GeNER can include relation extraction.
For instance, if we want to train a relation extraction model for the drug-disease relationship, we can simply ask ``\textit{Which drug is effective for disease?}'' and use the retrieved sentences as the positive training instances.
We can use retrieved phrases as objects (drugs) and leverage NER models to identify subject entities (diseases) in the evidence sentence.
It will be interesting to compare this approach with distantly supervised approaches~\citep{mintz-etal-2009-distant} in future research.
In addition, we discuss the potential of \ours~for fine-grained and zero-shot NER tasks in Appendices \ref{subsec:fine_grained} and \ref{sec:zeroshot}.

\section*{Limitations}

Until a superior QA model improves \ours~in the future, it will continue to inherit the limitations of the current QA model.
For instance, it is difficult to adapt our framework to languages with limited resources other than English because DensePhrases does not support other languages; moreover, other QA models are being primarily developed for English.
In this regard, we believe that future research on generating NER datasets for low-resource languages would be valuable and interesting.

Although our template ``\textit{Which} \typetoken\textit{?}'' can be generally applied to various \textit{named} entities, it requires modification for some entity types.
For instance, for extracting \textit{numerical} entity types such as money, date, time, duration, and quantity, questions beginning with ``when,'' ``what time,'' and ``how much/many,'' which are tailored to the specific types, would be required.
Our study focuses on named entities, and identifying these specialized entity types will be a part of of future research.

\section*{Acknowledgements}
We thank Jungsoo Park, Gyuwan Kim, Mujeen Sung, Sungdong Kim, Yonghwa Choi, Wonjin Yoon, and Gangwoo Kim for the helpful feedback.
This research was supported by (1) National Research Foundation of Korea (NRF-2020R1A2C3010638), (2) the MSIT (Ministry of Science and ICT), Korea, under the ICT Creative Consilience program (IITP-2021-2020-0-01819) supervised by the IITP (Institute for Information \& communications Technology Planning \& Evaluation), and (3) a grant of the Korea Health Technology R\&D Project through the Korea Health Industry Development Institute (KHIDI), funded by the Ministry of Health \& Welfare, Republic of Korea (grant number: HR20C0021). 

\bibliography{anthology,custom}
\bibliographystyle{acl_natbib}

\appendix

\clearpage
\appendix
\setcounter{table}{0}
\setcounter{figure}{0}
\renewcommand\thetable{\Alph{section}.\arabic{table}}
\renewcommand\thefigure{\Alph{section}.\arabic{figure}}

\section{Normalization Rules}
\label{appendix:rule_description}

This section details the normalization rules defined by us.
They are generally applicable to several domains without considerable modification.
They can be treated as hyperparameters (i.e., whether a rule is applied or not) and tuned based on the model performance on the validation sets.
They can also be determined without validation sets, based on the common knowledge of practitioners regarding target entities.
They must be applied in order for reproducing the same results.

\begin{itemize}
    \item \textbf{Rule 1} 
    Some retrieved phrases contain multiple entities linked by the conjunction ``\textit{and}.''
    We simply split such composite mentions based on ``\textit{and}.''
    This rule should be applied based on the annotation guidelines of the datasets.
    For instance, biomedical NER datasets such as NCBI-disease consider composite mentions (e.g., "\textit{colorectal, endometrial, and ovarian cancers}") as one entity;\footnote{\url{https://www.ncbi.nlm.nih.gov/CBBresearch/Dogan/DISEASE/Guidelines.html}} therefore, we do not use this rule for these datasets.
    Moreover, it should not be applied to {movie} entities (for example, ``\textit{Harry Potter and the Sorcerer's Stone}).''
    \item \textbf{Rule 2} 
    \densephrases~frequently returns phrases with punctuation at the start or end of the string, such as commas or quotation marks.
    We remove these noises using simple post-processing (e.g., \textit{Leprosy\textbf{,}} $\rightarrow$ \textit{Leprosy}).
    However, for some entities such as {songs}, punctuation may not be noise but can be a part of the name.
    \item \textbf{Rule 3} 
    We exclude phrases that are entirely in lowercase.
    Many named entities in the real world contain one or more uppercase letters (e.g., the first letter of a person's name is capitalized.). Thus, retrieved phrases, entirely in lowercase, are lowercase are more likely to be noisy results.
    However, because lowercase entities are common in biomedical datasets, this rule should be carefully applied depending on the datasets and entity types.
    \item \textbf{Rule 4} 
    We remove definite article ``\textit{the}'' from the string (e.g., ``\textit{the Boston Red Sox}'' $\rightarrow$ ``Boston Red Sox'').
    This rule should be applied depending on the annotation guidelines of the datasets or the superficial characteristics of the entities.
    For instance, because band entities sometimes include ``the'' in their name, this rule should not be applied to such entities.
    \item \textbf{Rule 5} 
    We exclude phrases with a length of less than three in our dictionary because short strings can cause significant noise. 
    \item \textbf{Rule 6} 
    We exclude phrases whose lowercase strings are in the stopword list such as ``\textit{WAS}'' (Wiskott-Aldrich Syndrome) and ``\textit{US}'' (United States) because they can cause considerable amount of false-positive noise in the dictionary matching process.
    Simultaneously, this rule can produce false-negative noise in the generated dataset; however, self-training mitigates this noise.
    \item \textbf{Rule 7} 
    We exclude phrases that are the same as \typetoken~in the sub-question. 
    For example, if we ask a question ``\textit{Which disease?},'' and the resulting phrase is ``\textit{disease},'' we do not use it.
    \item \textbf{Rule 8} 
    Because named entities are often abbreviated, it is important to annotate abbreviations so as to avoid false-negative noise from them. 
    We detected the abbreviations of retrieved phrases using the ScispaCy abbreviation tool.\footnote{\url{https://github.com/allenai/scispacy}}
    For instance, when the phrase ``\textit{Crohn's disease}'' is retrieved with the evidence sentence ``\textit{Crohn's disease (CD) is one of the two main forms of inflammatory bowel disease.},'' its abbreviation ``\textit{CD}'' is detected.
    It should be noted that abbreviations are not added to the dictionary because they usually have short forms, which can lead to considerable noise in dictionary matching.
    \item \textbf{Rule 9} 
    During dictionary matching, phrases in the dictionary and sentences are converted to lowercase by default.
    We prevent lowercase single tokens in the sentence from being matched with phrases in the dictionary because single lowercase tokens tend to be noisy (compared with multi-tokens).
    \item \textbf{Rule 10} 
    We use the phrase mining tool, AutoPhrase \cite{shang2018automated}, to refine entity boundaries in the dictionary matching stage.
    Specifically, if the span of a retrieved phrase is included in that of the phrase detected by AutoPhrase, we expand the span of the retrieved phrase to that of the detected phrase.
\end{itemize}


\setcounter{table}{0}

\section{Baseline Models}

\subsection{Low-resource NER}\label{appendix:low-baseline}
\paragraph{Seed Entities}
This model directly matches seed entities with the test corpus.
Seed entities are a small number of entities, which are manually selected by experts.
For instance, OSCE, NATO, Honda, Interfax, and Marseille are seed entities for the organization type of CoNLL-2003.

\paragraph{Neural Tagger}
We first annotate in-domain sentences by dictionary matching using the seed entity set as the dictionary, and then train \roberta~or \biobert~on the generated corpus.

\paragraph{Self-training}
Similar to our model, this model is trained through self-training on weak labels, but the labels are generated by dictionary matching between in-domain sentences and seed entities.
This can be viewed as a weak version of \bond~\cite{liang2020bond} that uses a small dictionary (i.e., the seed entity set).

\paragraph{\tallor} \cite{li-etal-2021-weakly}
This model is a strong baseline model that starts with 20-60 initial labeling rules (called seed rules) and automatically expands its labeling rule set.
Seed rules are defined as string matching between entity candidates (i.e., spans of text) in in-domain sentences and pre-defined seed entities.
A neural model that is initially trained on the sentences annotated by seed rules generates weak labels.
Rule candidates are selected based on the in-domain sentences and weak labels, and several top-ranked rules are added to the rule set.
This process is performed iteratively.

\paragraph{\bond} \cite{liang2020bond}
This framework first generates weak labels by dictionary matching between the in-domain sentences and in-domain dictionary, and then trains NER models based on self-training, which is the previous best weakly supervised method.
Dictionaries are created using rich external resources such as Wikidata and online websites.
\Cref{subsec:dictionary} provides more information on the in-domain dictionaries.
In the first iteration, the \textit{teacher} and \textit{student} models are initialized to standard language models (e.g. \roberta), and the teacher model is fine-tuned on the weak labels.
The teacher model then re-annotates the in-domain sentences, and the student model is trained on the newly generated labels by the teacher model.
For every $T_\text{update}$ iterations (i.e., the period of the update), the teacher model is updated as a (trained) student model.

\subsection{Few-shot NER}
\label{appendix:few_shot}

We used the models from the recent two studies~\cite{huang-etal-2021-shot,jia-etal-2022-question} as baselines.
We excluded some few-shot NER models because they use a sufficient amount of source data~\cite{yang-katiyar-2020-simple,cui-etal-2021-template}, which differs from our setting.

\paragraph{Supervised} 
This model is trained directly on few-shot examples.
\roberta~(for CoNLL-2003 and Wikigold) and \biobert~(for BC5CDR) were used in this study.

\paragraph{Noisy supervised pre-training (NSP)} \cite{huang-etal-2021-shot} 
NSP pre-trains models on the large-scale corpus WiNER \cite{ghaddar-langlais-2017-winer}, which comprises 2013 Wikipedia documents and weak labels for 113 fine-grained entity types.
The labels are generated based on the anchor links and coreference resolution.
The models pretrained by NSP were then fine-tuned using few-shot examples.

\paragraph{Self-training} This model~\cite{huang-etal-2021-shot} follows the current semi-supervised learning method~\cite{xie2020self}, where the model is initialized with few-shot examples and further (self-)trained using unlabeled training sentences.

\begin{table*}[t]
\centering
\footnotesize
\begin{adjustbox}{max width = 0.99\textwidth}

\begin{tabular}{llclc}
\toprule
\textbf{Dataset} & \textbf{Entity type} & \textbf{\# Seeds} & \textbf{Seed entities}  \\
\midrule
\multirow{4}{*}{CoNLL-2003} & \textit{person} & 7 & \begin{tabular}[c]{@{}l@{}} 
wasim akram, waqar younis, mushraq ahmed, aamir sohail, saeed anwar, bill clinton, \\mother teresa
\end{tabular}  \\
\cmidrule{2-4}
& \textit{location} & 8 & \begin{tabular}[c]{@{}l@{}}
britain, italy, russia, sweden, belgium, iraq, south africa, united states
\end{tabular}  \\
\cmidrule{2-4}
& \textit{organization} & 5 & \begin{tabular}[c]{@{}l@{}}
osce, nato, honda, interfax, marseille
\end{tabular}   \\
\midrule
\multirow{6}{*}{Wikigold} & \textit{person} & 5 & cabral, bobick, belgrano, behe, moses mendelssohn   \\
\cmidrule{2-4} 
& \textit{location} & 10 & 
\begin{tabular}[c]{@{}l@{}} 
maaa, ncaa, 139th, major league baseball, cbs cable, bcit, montreal hockey club, 882 6pr, konami, \\
30 seconds to mars
\end{tabular}   \\
\cmidrule{2-4}
& \textit{organization} & 10 & 
\begin{tabular}[c]{@{}l@{}} 
england, indonesia, old goa, chicago, ontario, aabenraa county, illinois, hay street, b \& sr, \\ cal anderson park
\end{tabular}  \\
\midrule
\multirow{14}{*}{WNUT-16} & \textit{person} & 2 & lindsay lohan, scooter braun    \\
\cmidrule{2-4}
& \textit{location} & 3 & belgium, toronto, arizona  \\
\cmidrule{2-4}
& \textit{product} & 8 & \begin{tabular}[c]{@{}l@{}} ipad, htc desire z, iphone, pumpkin moonshine, coke, flip minohd, club penguin, xbox 360 \end{tabular}  \\
\cmidrule{2-4}
& \textit{facility} & 5 & \begin{tabular}[c]{@{}l@{}}
visions lounge, frat house hattiesburg, empire state building, disney world, club blu \end{tabular}  \\
\cmidrule{2-4}
& \textit{company} & 3 & twitter, youtube, facebook  \\
\cmidrule{2-4}
& \textit{sports team} & 1 & jv soccer  \\
\cmidrule{2-4}
& \textit{TV show} & 1 & friday night lights  \\
\cmidrule{2-4}
& \textit{movie} & 1 & iron man 2  \\
\cmidrule{2-4}
& \textit{music artist} & 1 & kings of leon  \\
\midrule
\multirow{1}{*}{NCBI-disease} & \textit{disease} & 20 & \begin{tabular}[c]{@{}l@{}} 
dmd, pws, myotonic dystrophy, g6pd deficiency, hd, pku, aniridia, duchenne muscular dystrophy, \\
fap, a - t, tay - sachs disease, tsd, fmf, prader - willi syndrome, amn, wiskott - aldrich syndrome, \\ huntington disease, pelizaeus - merzbacher disease, bmd 
\end{tabular}     \\
\midrule
\multirow{3}{*}{BC5CDR} & \textit{disease} & 10 & \begin{tabular}[c]{@{}l@{}}proteinuria, esrd, thrombosis, tremor, hepatotoxicity, hypertensive, thrombotic microangiopathy, \\ thrombocytopenia, akathisia, confusion \end{tabular}   \\
\cmidrule{2-4}
& \textit{chemical} & 10 & \begin{tabular}[c]{@{}l@{}}nicotine, morphine, haloperidol, warfarin, clonidine, creatinine, isoproterenol, \\ cyclophosphamide, sirolimus, tacrolimus\end{tabular}  \\
\midrule
\multirow{1}{*}{CHEMDNER} & \textit{chemical} & 60 & \begin{tabular}[c]{@{}l@{}}
glucose, cholesterol, glutathione, ethanol, androgen, graphene, glutamate, dopamine, cocaine, \\
serotonin, estrogen, nicotine, tyrosine, resveratrol, nitric oxide, cisplatin, alcohol, superoxide, \\ curcumin, metformin, amino acid, testosterone, flavonoids, camp, methanol, amino acids, \\
fatty acids, polyphenols, nmda, silica, 5-ht, oxygen, calcium, copper, cadmium, arsenic, zinc, \\
mercury, (1) h, ca (2+) 
\end{tabular}   \\
\bottomrule
\end{tabular}
\end{adjustbox}
\caption{
Seed entities used in our experiments.
All seed entities are in lowercase.
We use the seed entities provided by \citet{li-etal-2021-weakly} for CoNLL-2003, BC5CDR, and CHEMDNER.
For the remaining datasets where seed entities are not provided, we manually select frequent and high-precision entities, following \citet{li-etal-2021-weakly}.
}
\label{tab:seed_entities}
\end{table*}

\paragraph{\quip~(standard)} \cite{jia-etal-2022-question}
\quip~is a contextualized representation model pre-trained with 80 million question-answer pairs, which are automatically generated by the BART-large model~\cite{lewis-etal-2020-bart}.
\quip~(standard) comprises a \quip~encoder with a randomly initialized linear output layer. The initialized \quip~model is fine-tuned on few-shot examples.

\paragraph{\quip~(Q-prompt)} \cite{jia-etal-2022-question}
Unlike \quip~(Standard), the output layer of \quip~(Q-prompt) is initialized as the embeddings for question prompts.
For instance, for the \textit{organization} type, ``\textit{What is an organization?}'' is used as the question prompt, and the output layer is then initialized as \quip's representation of the prompt.
\citet{jia-etal-2022-question} showed that this initialization strategy is effective for few-shot NER.
As suggested, we used the same question prompts as~\citet{jia-etal-2022-question} for CoNLL-2003 and Wikigold.
For BC5CDR, we used ``\textit{What is a disease?}'' for the disease type and ``\textit{What is a chemical compound?} for the chemical type'' because ``\textit{What is a drug?}'' is less effective.

\setcounter{table}{0}
\section{Implementation Details}
\label{appendix:implementation_details}

\paragraph{Seed entities}
\label{subsec:seed_entities}
For CoNLL-2003, BC5CDR, and CHEMDNER, we used the same seed entities as \citet{li-etal-2021-weakly}.
For the other benchmarks, because there are no pre-defined seed entities, we manually selected frequent and high-precision entities in the training sets, following \citet{li-etal-2021-weakly}.
Note that this selection process relies on in-domain resources, such as training sentences or professional knowledge of experts.
The seed entities are listed in \Cref{tab:seed_entities}.

\paragraph{In-domain dictionary}
\label{subsec:dictionary}
\bond~requires rich in-domain dictionaries to achieve a high performance.
For CoNLL-2003, Wikigold, and WNUT-16, the model used Wikidata that comprises more than 96 million entities (as of October 21, 2021) and multiple gazetteers from different websites for each dataset.
For instance, 10 websites were used for CoNLL-2003, including Random Name and Intergovernmental Organization.\footnote{See~\citet{liang2020bond} for the entire list and website links.}
For NCBI-disease and BC5CDR, we used dictionaries provided by \citet{shang-etal-2018-learning}, which are derived from the MeSH database and the Comparative Toxicogenomics Database, which comprises more than 300k disease and chemical entities.
The dictionaries were additionally tailored to the target corpora using techniques such as corpus-aware dictionary tailoring~\cite{shang-etal-2018-learning}.

\clearpage

\begin{table*}[t!]
\centering
\footnotesize
\begin{adjustbox}{max width = \textwidth}
\begin{tabular}{lllcc}
\toprule
\textbf{Dataset} & \textbf{Entity Type} & \typetoken & \boldsymbol{$k_l$} & \textbf{Rule}   \\
\midrule
\multirow{4}{*}{CoNLL-2003} & \multirow{1}{*}{\textit{person}} &   athlete, politician, actor  & \multirow{1}{*}{5,000} & \multirow{1}{*}{1,3,4} \\
\cmidrule{2-5}
& \multirow{1}{*}{\textit{location}} &   country, city, state in the USA  & \multirow{1}{*}{5,000} & \multirow{1}{*}{1,3,4}  \\
\cmidrule{2-5}
& \multirow{1}{*}{\textit{organization}} &   sports team, company, institution  & \multirow{1}{*}{5,000} & \multirow{1}{*}{1,3,4}  \\
\midrule
\multirow{4}{*}{Wikigold} & \multirow{1}{*}{\textit{person}} & athlete, politician, actor, director, musician  & \multirow{1}{*}{4,000} & \multirow{1}{*}{1,3,4} \\
\cmidrule{2-5}
& \multirow{1}{*}{\textit{location}} &   country, city, state in the USA, road, island  & \multirow{1}{*}{4,000} & \multirow{1}{*}{1,3,4}  \\
\cmidrule{2-5}
& \multirow{1}{*}{\textit{organization}} &  sports team, company, institution, association, band  & \multirow{1}{*}{4,000} & \multirow{1}{*}{1,3,4}  \\
\midrule
\multirow{16}{*}{WNUT-16} & \multirow{1}{*}{\textit{person}} &  athlete, politician, actor, author  & \multirow{1}{*}{1,000} & \multirow{1}{*}{1,3,4}  \\
\cmidrule{2-5}
& \multirow{1}{*}{\textit{location}} &   country, city, state in the USA  & \multirow{1}{*}{1,000} & \multirow{1}{*}{1,3,4}  \\
\cmidrule{2-5}
& \multirow{2}{*}{\textit{product}} &  mobile app  & \multirow{1}{*}{1,000} & 3 \\
& &   software, operating system, car, smart phone  & 1,000 & 1,3,4 \\
\cmidrule{2-5}
& \multirow{2}{*}{\textit{facility}} &   facility, cafe, restaurant, college, music venue  & \multirow{1}{*}{1,000} & 3 \\
& &   sports facility  & 1,000 & 1,3,4 \\
\cmidrule{2-5}
& \multirow{2}{*}{\textit{company}} & company, technology company & 1,000 & 1,3,4 \\
& & news agency, magazine  & 1,000 & 1,3 \\
\cmidrule{2-5}
& \multirow{1}{*}{\textit{sports team}} &   sports team  & 1,000 & 1,3,4  \\
\cmidrule{2-5}
& \multirow{1}{*}{\textit{TV show}} & TV show  & 1,000 & 3  \\
\cmidrule{2-5}
& \multirow{1}{*}{\textit{movie}} &   movie  & 1,000 & 3 \\
\cmidrule{2-5}
& \multirow{1}{*}{\textit{music artist}} & band, rapper, musician, singer & \multirow{1}{*}{1,000} & \multirow{1}{*}{3} \\
\midrule
\multirow{1}{*}{NCBI-disease} & \multirow{1}{*}{\textit{disease}} & disease  & 35,000 & 4,9 \\ 
\midrule
\multirow{3}{*}{BC5CDR} & \multirow{1}{*}{\textit{disease}} & disease  & 15,000 & 4,9 \\
\cmidrule{2-5}
& \multirow{1}{*}{\textit{chemical}} & chemical compound, drug & 15,000 & 4,9 \\
\midrule
\multirow{1}{*}{CHEMDNER} & \multirow{1}{*}{\textit{chemical}} & chemical compound, drug & 10,000 & 4,9 \\
\midrule
\multirow{5}{*}{CrossNER} & \textit{enzyme} &  enzyme & 5,000 & 1,4,9  \\
\cmidrule{2-5}
& \textit{astronomical object} & astronomical object  & 5,000 & 1,3,4  \\
\cmidrule{2-5}
& \textit{award} & award  & 10,000 & 1,3,4  \\
\cmidrule{2-5}
& \textit{conference} & conference on artificial intelligence & 5,000 & 3 \\
\bottomrule
\end{tabular}
\end{adjustbox}
\caption{
Subquestions and hyperparameters used for NER benchmarks.
Each sub-question is formulated as ``\textit{Which} \typetoken\textit{?}'' and used for the retrieval.
$k_l$: number of unique sentences retrieved for each sub-question.
The total number of sentences for a dataset is calculated as the sum of the number of sentences for each sub-question.
Normalization rules are detailed in~\Cref{appendix:rule_description}.
Note that we omit Rules \textsc{2}, \textsc{5}, \textsc{6}, \textsc{7}, \textsc{8}, and \textsc{10}, because they are commonly applied.
}
\label{tab:discrete_config}
\vspace{-3mm}
\end{table*}

\clearpage

\paragraph{\tallor}
Although we used the official code base,\footnote{\href{https://github.com/JiachengLi1995/TALLOR}{https://github.com/JiachengLi1995/TALLOR}} the model we implemented was lower than the reported performance (scores with $^\dagger$ in \Cref{tab:main_table}).
This is because, in the original implementation, n-gram statistics of the test set was used in the entity candidate generation process of \tallor.
On the other hand, we implemented the model using only the training corpus for fair comparison.

\paragraph{Other details}
The entire subquestions we selected are listed in \Cref{tab:discrete_config}.
We used public PyTorch implementation provided by \citet{liu2019roberta} and \citet{lee2020biobert}\footnote{\href{https://github.com/dmis-lab/biobert-pytorch}{https://github.com/dmis-lab/biobert-pytorch}} for implementing the Neural Tagger baselines and our fine-tuning models.
We used the pre-trained weights of the \texttt{densephrases-multi-query-multi} model for the question and phrase encoders of DensePhrases.\footnote{\href{https://github.com/princeton-nlp/DensePhrases}{https://github.com/princeton-nlp/DensePhrases}}
For \bond, we used the official code base provided by the authors.\footnote{\href{https://github.com/cliang1453/BOND}{https://github.com/cliang1453/BOND}}
Also, we used the same code base for the self-training step in \ours.
The best hyperparameters in self-training of our models are detailed in \Cref{tab:self_training_config}.

\begin{table}[t!]
\centering
\footnotesize
\begin{tabular}{clcc}
\toprule
\textbf{Model} & \textbf{Dataset} & \boldsymbol{$T_\text{begin}$} & \boldsymbol{$T_\text{update}$}  \\
\midrule
\multirow{11}{*}{\ours} & CoNLL-2003 & 900 & 300  \\
& Wikigold & 500 & 300  \\
& WNUT-16 & 900 & 450  \\
\cmidrule{2-4}
& NCBI-disease & 900 & 300  \\
& BC5CDR & 500 & 200  \\
& CHEMDNER & 900 & 300  \\
\cmidrule{2-4}
& Enzyme & 350 & 700  \\
& Astr. & 500 & 300  \\
& Award & 350 & 400  \\
& Conf. & 200 & 100  \\
\midrule
\multirow{5}{*}{\bond} & CoNLL-2003 & 900 & 450 \\
 & Wikigold & 900 & 300 \\
 & WNUT-16 & 900 & 300 \\
 \cmidrule{2-4}
 & NCBI-disease & 900 & 450 \\
 & BC5CDR & 500 & 300 \\
 \midrule
\multirow{6}{*}{Self-training} & CoNLL-2003 & 400 & 100 \\
 & Wikigold & 350 & 200 \\
 & WNUT-16 & 500 & 100 \\
 \cmidrule{2-4}
 & NCBI-disease & 200 & 100 \\
 & BC5CDR & 500 & 100 \\
 & CHEMDNER & 900 & 450 \\
\bottomrule
\end{tabular}
\caption{
Hyperparameter configuration in self-training of \ours~and baselines. 
$T_\text{begin}$ is the early stopping step before updating the model, and $T_\text{update}$ is the period of the update.
For more detailed descriptions of $T_\text{begin}$ and $T_\text{update}$, refer to \citet{liang2020bond}.
}
\label{tab:self_training_config}
\end{table}

\setcounter{table}{0}
\setcounter{figure}{0}

\section{Fine-grained NER}
\label{subsec:fine_grained}

\begin{table}[t!]
\centering
\footnotesize

\begin{adjustbox}{max width = 0.49\textwidth}

\begin{tabular}{lcc}
\toprule
\textbf{Dataset} & \textbf{\begin{tabular}[c]{@{}c@{}} \textbf{\# Sents} \\ (\textbf{train / valid / test}) \end{tabular}} & \textbf{\begin{tabular}[c]{@{}c@{}} \textbf{\# Labels} \\ (\textbf{train / valid / test}) \end{tabular}} \\
\midrule
Enzyme & 200 / 450 / 543 & 22 / 48 / 80 \\
Astro. & 200 / 450 / 543 & 121 / 373 / 337 \\
\midrule
Award & 100 / 400 / 416 & 34 / 124 / 141 \\
\midrule
Conference & 100 / 350 / 431 & 24 / 89 / 93 \\
\bottomrule
\end{tabular}

\end{adjustbox}

\caption{
Statistics of fine-grained NER datasets derived from the CrossNER dataset~\cite{liu2021crossner}.
Astr.: astronomical object.
\#~Types: number of entity types. 
\#~Sents: number of sentences.\
\#~Labels: number of entity-level annotations.
}
\label{tab:statistics_crossner}
\end{table}
\begin{table}[t!]
\footnotesize
\centering
\begin{tabularx}{0.49\textwidth}{ lc *{4}{Y} }
\toprule
\textbf{Model}  & \textbf{Enzyme} & \textbf{Astr.} & \textbf{Award} & \textbf{Conf.} \\
\midrule
Fully supervised & 56.4 & 78.0 & 75.4  & 49.4  \\
\midrule
\ours  & 49.5 & 71.9 & 80.9 & 41.1 \\
\quad+ Fine-tuning & \textbf{63.1} & \textbf{86.8} & \textbf{81.6} & \textbf{64.0} \\
\bottomrule
\end{tabularx}
\caption{
Performance of the fully supervised \roberta~model and \ours~on fine-grained entity types.
\fscore~score is reported.
Astr. and Conf. indicate astronomical object and conference, respectively.
}
\label{tab:fine_grained}
\end{table}

The human annotations for fine-grained entity types are sparser than those for coarse-grained types.
To determine whether the data sparsity problem is addressable by \ours, we created four fine-grained datasets derived from CrossNER~\cite{liu2021crossner}: {enzyme} and {astronomical object} (natural science domain), {award} (literature domain), and {conference} (artificial intelligence domain), by removing labels for the other entity types from the sentences in CrossNER.
We selected these four types because they were not coarse-grained, thus meeting the purpose of this experiment.
\Cref{tab:statistics_crossner} presents the statistics of the datasets.
We used a single sub-question for each dataset (See \Cref{tab:discrete_config}.)

\paragraph{Results}
We compared \ours~with the fully supervised \roberta, as shown in Table \ref{tab:fine_grained}.
We found that \ours~is highly comparable with the fully supervised model and sometimes even outperforms the fully supervised model (Award).
Its performance can be further improved by fine-tuning on each small training set (+ Fine-tuning).

\begin{table*}[t]
\centering
\footnotesize

\begin{adjustbox}{max width = 0.99\textwidth}
\begin{tabular}{llcc}
\toprule
\typetoken & \textbf{Retrieved Entities from \ours} & \textbf{P@50} & \textbf{Diversity} \\
\midrule
song nominated for the Grammy Awards & \begin{tabular}[c]{@{}l@{}}Hotline Bling, Love Me like You Do, Mystery of Love, \\ Can't Stop the Feeling!, The Price is Wrong, ... \end{tabular} & 0.96 & 0.90 \\
\midrule
dish made with eggs & \begin{tabular}[c]{@{}l@{}}Eggs Benedict, Pancakes, Shakshouka, Omelettes, \\ Huevos rancheros, Chilaquiles, Menemen, ... \end{tabular}  & 0.80 & 0.78  \\
\midrule
satellite made by an American company & \begin{tabular}[c]{@{}l@{}} GE-2, AMC-2, Ariel 1, Syncom 3, CHIPSat, \\ Telstar, Explorer 1, Westar 1, SkyTerra-1, ... \end{tabular} & 0.88 & 0.82 \\
\bottomrule
\end{tabular}
\end{adjustbox}
\caption{
Retrieval entities from \ours~for fine-grained entity types.
``\textit{Which} \typetoken\textit{?}'' is used as a question.
}
\label{tab:fine_grained_2}
\end{table*}

\paragraph{Retrieved entities}
We further show the potential of \ours~on three entities that are extremely fine-grained such as ``\textit{satellite made by an American company}.''
As there are no human annotations for these entities, we manually measured retrieval performance using precision at 50 (P@50) and diversity, similar to Section \ref{subsec:query_template_selection}.
\Cref{tab:fine_grained_2} shows that accurate and diverse entities were retrieved for each question.
Because of the flexibility of the natural language questions, \ours~can easily provide NER models for specialized entity types.

\setcounter{table}{0}

\section{Connection to Zero-shot NER}
\label{sec:zeroshot}
Zero-shot NER aims to build models that generalize to \textit{unseen} entity types without corresponding labels.
It has hard constraints that the entity types in $\mathbf{Y}_\text{test}$ are not observed during training over $\mathcal{D}_\text{train}$. 
To tackle this task, researchers have proposed to utilize external descriptions of the target entities~\citep{aly-etal-2021-leveraging,wang-etal-2021-learning-language-description}.
We expect that \ours~can support zero-shot NER by generating weak-labeled sentences for the target entity types, where the sentences can be used as semantic information.
Leveraging GeNER in zero-shot NER would be interesting since it can remove the strong assumption that type descriptions are available.

\end{document}